\pdfoutput=1

\documentclass[11pt]{article}

\usepackage[preprint]{acl}

\usepackage{times}
\usepackage{latexsym}

\usepackage[T1]{fontenc}

\usepackage[utf8]{inputenc}

\usepackage{microtype}

\usepackage{inconsolata}

\usepackage{graphicx}

\usepackage{amsfonts}
\usepackage{mathtools}

\usepackage{algorithm}
\usepackage{algorithmicx}
\usepackage{algcompatible}
\usepackage{xcolor}
\usepackage{multirow}
\usepackage{colortbl}
\usepackage{adjustbox}
\usepackage{tabularx, booktabs}
\usepackage{caption}
\usepackage{subcaption}
\usepackage{tcolorbox}
\usepackage{enumitem}

\usepackage{pifont}
\newcommand{\cmark}{\ding{51}}
\newcommand{\xmark}{\ding{55}}
\newcommand{\sftype}[1]{{\textsf{\small #1}}}

\usepackage{amsmath}
\usepackage{xspace}
\usepackage{makecell}
\usepackage{subcaption}

\usepackage{amsthm}
\newtheorem{theorem}{Theorem}
\usepackage{graphicx}
\usepackage{subcaption}
\usepackage{float}
\title{Efficient Latent Semantic Clustering \\ for Scaling Test-Time Computation of LLMs}

\author{
 \textbf{Sungjae Lee\textsuperscript{1}},
 \textbf{Hoyoung Kim\textsuperscript{2}},
 \textbf{Jeongyeon Hwang\textsuperscript{2}},
 \textbf{Eunhyeok Park\textsuperscript{1,2}},
 \textbf{Jungseul Ok\thanks{Corresponding Author}\textsuperscript{1,2}},
\\
\\
 \textsuperscript{1}Department of Computer Science and Engineering, POSTECH, South Korea \\
 \textsuperscript{2}Graduate School of Artificial Intelligence, POSTECH, South Korea
\\
{\texttt{\{sungjaelee25,cskhy16,jeongyeon.hwang,eh.park,jungseul\}@postech.ac.kr}}
 }

\begin{document}
\maketitle

\begin{abstract}
Scaling test-time computation—generating and analyzing multiple or sequential outputs for a single input—has become a promising strategy for improving the reliability and quality of large language models (LLMs), as evidenced by advances in uncertainty quantification and multi-step reasoning. A key shared component is semantic clustering, which groups outputs that differ in form but convey the same meaning. Semantic clustering enables estimation of the distribution over the semantics of outputs and helps avoid redundant exploration of reasoning paths. However, existing approaches typically rely on external models, which introduce substantial computational overhead and often fail to capture context-aware semantics. We propose Latent Semantic Clustering (LSC), a lightweight and context-sensitive method that leverages the generator LLM’s internal hidden states for clustering, eliminating the need for external models. Our extensive experiment across various LLMs and datasets shows that LSC significantly improves the computational efficiency of test-time scaling while maintaining or exceeding the performance of existing methods.

\end{abstract}

\section{Introduction}

Scaling test-time computation has emerged as a promising strategy to improve the reliability and quality of responses from large language models (LLMs) by generating multiple responses for a single input \cite{kuhn2023semantic, lin2024generating, nikitin2024kernel, yao2023tree, hao2023reasoning, snell2025scaling}. For reliability, uncertainty quantification methods assess the semantic diversity among the multiple outputs to estimate model confidence~\cite{kuhn2023semantic, lin2024generating, nikitin2024kernel}. For quality, reasoning methods explore and aggregate multiple reasoning paths \cite{yao2023tree, hao2023reasoning}.

\begin{figure}[t!]
    \centering
    \includegraphics[width=0.48\textwidth]{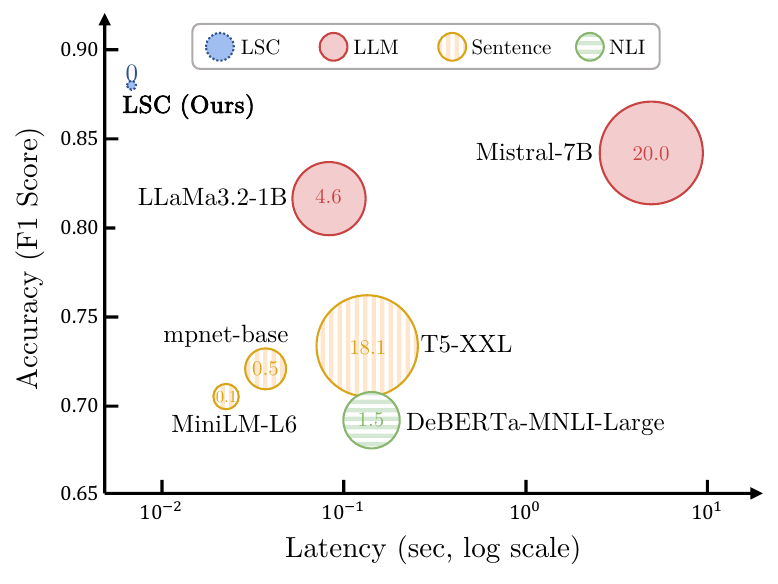}
    \vspace{-5mm}
    \caption{Comparison of LSC and NLI/embedding models in terms of latency (sec), clustering performance (F1 score), and memory usage (GB). F1 score is plotted against log-scaled latency; circle size denotes square-root memory usage. Circles with the same hatch pattern denote the same model type (i.e., LLM, sentence-embedding, or NLI) and numbers inside circles indicate memory usage. LSC achieves the best performance with minimal latency and memory usage. Detailed experimental results are described in Table~\ref{tab:clustering-perform}.}
    \label{fig:scatter}
    \vspace{-3mm}
\end{figure}

A key component in such test-time scaling methods is {\it semantic clustering}, which groups responses with the same meaning (despite their diverse forms). Indeed, effective clustering can improve uncertainty quantification \cite{kuhn2023semantic} by computing the distribution over distinct meanings of responses, and accelerate multi-step reasoning by avoiding unnecessary explorations of (semantically) duplicated reasoning paths \cite{lee2025semantic}. 
However, semantic clustering is challenging in open-ended tasks, where the output space is large, ambiguous, and diverse. Reasoning tasks, in particular, are inherently open-ended.

For semantic clustering, existing test-time scaling methods \cite{kuhn2023semantic, qiu2024semantic} typically employ external models such as natural language inference (NLI) \cite{nikitin2024kernel, lin2024generating} or sentence embedding models \cite{abdaljalil2025sindex}. Such approaches with external models incur significant computational overhead due to additional inference. In addition, they often fail to capture context-dependent semantics, as they operate separately from the generator of responses. As a result, they are both inefficient and less applicable to test-time scaling methods.

\begin{figure}[t!]
    \centering
    \includegraphics[width=0.49\textwidth]{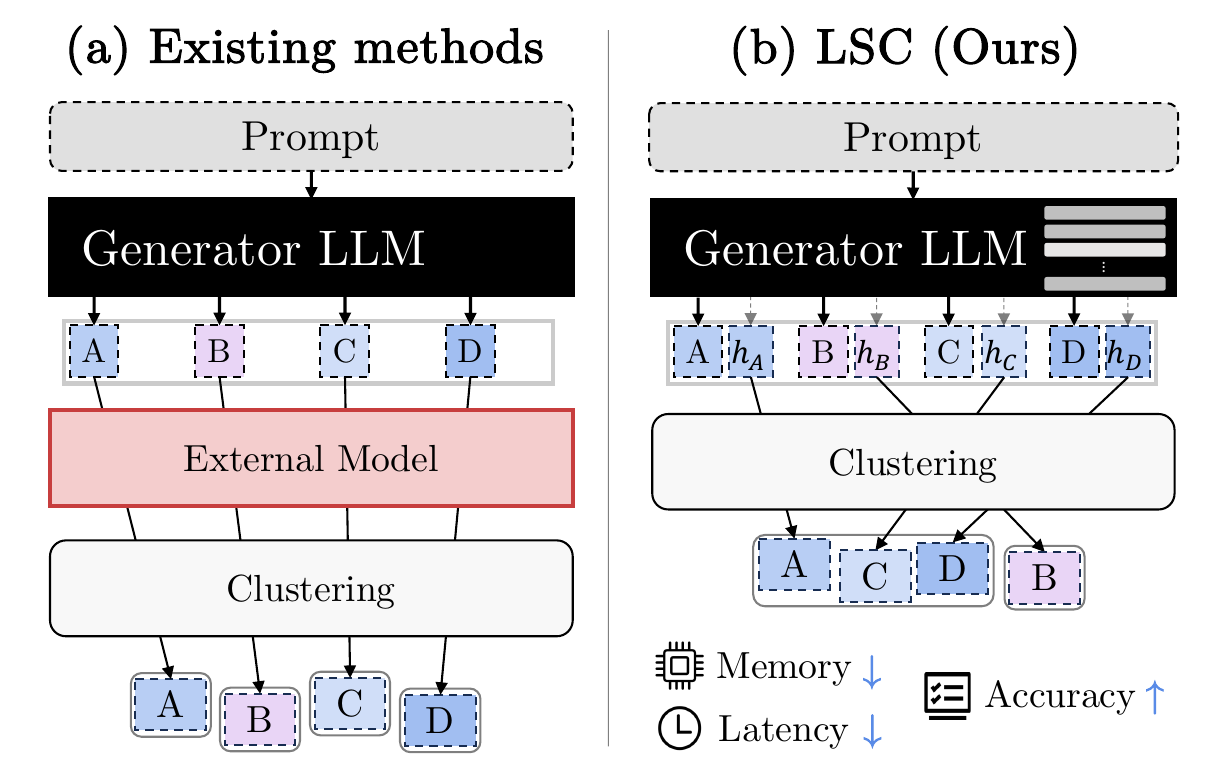}
    \caption{Comparison of semantic clustering frameworks with test-time computation scaling scenario with LLMs. (a) Previous methods rely on external NLI/embedding models to cluster generated sequences. (b) Our latent semantic clustering offers a more efficient and effective alternative by directly leveraging internal hidden representations of LLMs. same colors (\textit{i.e.}, blue and purple) denote semantically equivalent meanings. A concrete example with generated sequences by LLMs is shown in Figure~\ref{fig:se-case}.}
    \label{fig:pipeline}
\end{figure}

To address these limitations, we propose 
a lightweight yet effective method for semantic clustering, called Latent Semantic Clustering (LSC). The key idea is to directly leverage the internal latent representations of the generator, eliminating the need for external models (Figure~\ref{fig:pipeline}). LSC offers two main advantages for test-time scaling as shown in Figure~\ref{fig:scatter}. First, it is highly {\it efficient}—requiring only minimal memory to store hidden vectors and incurring virtually no additional computation. Second, it naturally captures context-aware semantics embedded in the generator’s representations, leading to more accurate clustering and effective improvement of test-time scaling methods.
Our experiment across various LLMs and datasets validates the efficiency and effectiveness of LSC in uncertainty quantification and multi-step reasoning.

\section{Background: Semantic Clustering}
This section begins with a formal description of semantic clustering (Section~\ref{sec:sc}).
We then describe how semantic clustering can be applied to enhance uncertainty quantification (Section~\ref{sec:su}) and to facilitate efficient exploration in LLM reasoning (Section~\ref{sec:se})—both of which commonly leverage test-time computation scaling. The related work is discussed in detail in Appendix~\ref{append:related-work}.

\subsection{Semantic Clustering with LLMs}\label{sec:sc}
To identify semantically consistent or distinct sequences generated by LLMs under test-time scaling strategies, semantic clustering groups outputs based on their semantics rather than lexical form. 
Specifically, we generate a set of sequences $\mathcal{S} = \{\mathbf{s}_1, \dots, \mathbf{s}_N\}$ by sampling from the generator LLM $\mathcal{M}$ given a context $x$.
Here, the context $x$ may include task instructions, in-context examples, the input problem or question, and—when applicable—intermediate reasoning trajectories from earlier steps.
To identify the semantic relationships among the generated sequences, the most widely used approach~\cite{kuhn2023semantic} employ bi-directional classification using an NLI model. 
It evaluates both $\theta_{ext}(\mathbf{s}_i,\mathbf{s}_j)$ and $\theta_{ext}(\mathbf{s}_j,\mathbf{s}_i)$, where $\theta_{ext}$ denotes an external model that determines whether one sequence semantically entails the other. 
If both classification results are \textit{entailment}, the two sequences are grouped into a semantic cluster $c \in \mathcal{C}$, which is called bi-directional entailment clustering (BDEC) proposed for LLM uncertainty quantification.
Recently, subsequent works~\cite{abdaljalil2025sindex} have replaced $\theta_{ext}$ with embedding models such as BERT~\cite{devlin2019bert} and Sentence-BERT~\cite{reimers-gurevych-2019-sentence}, applying clustering algorithms to the embeddings.

\subsection{Semantic Uncertainty}\label{sec:su}
In open-ended natural language tasks, semantic clustering can be utilized to measure the uncertainty in the $N$ texts generated by an LLM for a given context $x$.
Let $\mathcal{C}$ be the set of semantic clusters over the generated sequences $\mathbf{s}_1, ..., \mathbf{s}_N$.
The cluster probability $p(c \mid x, \mathcal{S})$ is defined by summing the probabilities of sequences assigned to cluster $c \in \mathcal{C}$ as follows:
\begin{align}
p(c \mid x, \mathcal{S}) := \sum_{\mathbf{s} \in \mathcal{S}} 1[\mathbf{s} \in c] \cdot p(\mathbf{s} \mid x) \;,
\end{align}
where $p(\mathbf{s}  \mid x)$ is either computed from token-level generation probabilities as $\prod_i p(s^i \mid s^{<i}, x)$ or approximated as $1/N$. 

Based on this distribution, the semantic entropy~\cite{kuhn2023semantic} on generated sequences $\mathcal{S}$ is defined as:
\begin{align}
\!\!\!\! \text{SE}(\mathcal{S}; x, \mathcal{C}) \!\!:=\! -\! \sum_{c \in \mathcal{C}} p(c \mid x, \mathcal{S}) \log p(c \mid x, \mathcal{S}) \!\!\;. \!\!
\end{align}
When generator LLMs produce diverse meanings from a single prompt, semantic entropy is high; when meanings converge, entropy is low, serving as a measure of LLM uncertainty.
Here, the reliability of semantic uncertainty estimation hinges on the quality of the clusters $\mathcal{C}$, which are typically derived using external models.

\begin{figure}[t!]
    \centering
    \includegraphics[width=0.49\textwidth]{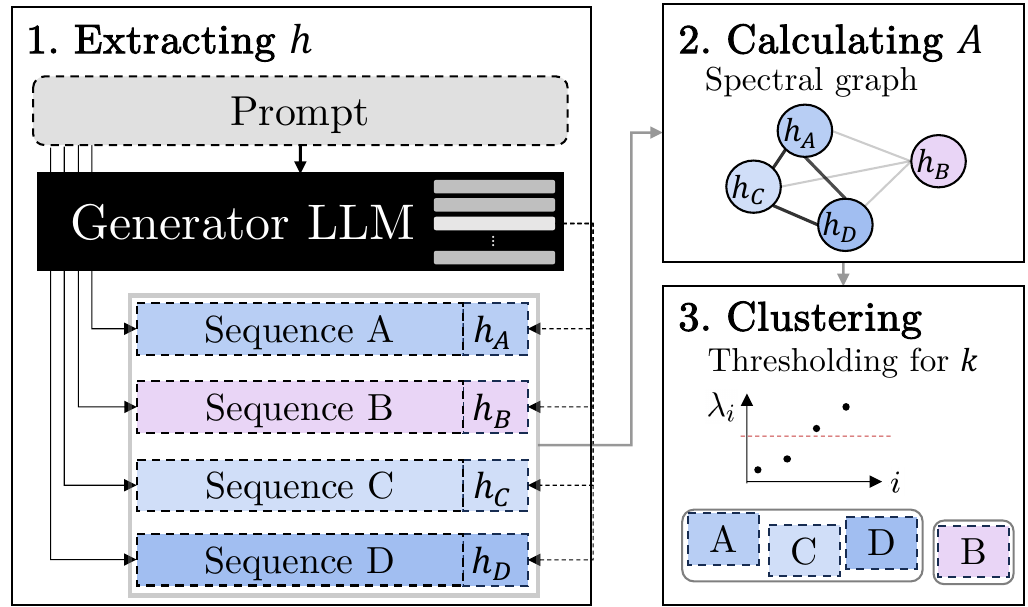}
    \caption{Overview of Latent Semantic Clustering (LSC). LSC consists of three main steps: (1) Extract hidden states during generations, (2) compute pairwise similarities to form an adjacency matrix, and (3) apply spectral clustering while determining the number of clusters $k$.}\label{fig:algorithm}
    \vspace{-3mm}
\end{figure}

\subsection{Semantic Exploration}\label{sec:se}
Beyond linear reasoning methods~\cite{wei2022chain, wang2022self}, tree search-based methods expand and explore intermediate reasoning trajectories to enhance the reasoning capabilities of LLMs~\cite{hao2023reasoning,yao2023tree} as test-time scaling strategies.
However, these methods often incur significant computational overhead due to redundant exploration of semantically equivalent paths. 
To alleviate this, recent studies have leveraged semantic clustering to reduce the number of expansions from all $d$ candidate nodes generated by the LLM at each step to only $d' < d$ semantically distinct nodes, thereby enabling more efficient exploration on reasoning trees~\cite{lee2025semantic,wang2025don}. 

Nonetheless, these methods still rely on external models while demanding additional computational resources and often struggle to effectively incorporate contextual information during reasoning.
Notably, \citet{lee2025semantic} constructed the input of NLI model using only the generated sequences without context, while \citet{wang2025don} performed additional training on external embedding model due to its limited capacity to effectively handle contextual information as described in Figure~\ref{fig:se-case} and Table~\ref{tab:clustering-perform}.
In contrast, we perform clustering without external resources by directly utilizing the internal semantic representations of the generator LLMs.

\section{Method}\label{sec:method}

For efficient semantic clustering, we propose Latent Semantic Clustering (LSC) that captures context-aware semantics by leveraging the hidden states produced by the generator LLMs.
Our approach involves three key steps: 
(1) we extract the hidden states from the generator LLM during inference; (2) we construct an adjacency matrix that maps pairwise semantic similarity between the generated sequences; (3) we perform spectral clustering while determining the optimal number of clusters. The overall procedure is summarized in Figure~\ref{fig:algorithm} and Algorithm~\ref{alg1}.

\begin{algorithm}[t]
\caption{Latent Semantic Clustering}
\begin{algorithmic}[1]
\REQUIRE \!\! Input $x$, \!\! LLM $\mathcal{M}$, \!\! threshold $\tau$
\ENSURE Set of clusters $\mathcal{C}$
\STATE During generation of $N$ sequences $\mathcal{S}$ with $\mathcal{M}(x)$, obtain hidden states $h_n$ for each $s \in \mathcal{S}$
\STATE Construct adjacency matrix $A$ using cosine similarity between hidden states via~\eqref{eq:cosine-sim}
\STATE Determine the number of clusters $k$ by thresholding the eigenvalues of~\eqref{eq:laplacian} at $\tau$
\STATE Obtain the cluster set $\mathcal{C}$ of size $k$ by applying spectral clustering on $A$
\end{algorithmic}
\label{alg1}
\end{algorithm}

\paragraph{Extracting hidden states} 
The first step of LSC extracts latent representations that capture the semantic content of each generated sequence.  
To do so, we leverage the hidden states from an intermediate Transformer layer during the auto-regressive generation process for $i$-th token, defined as:
\begin{align}
h^{i, 0} &= e^i + p^i \\
h^{i, \ell} &= \text{TransformerBlock}^{(\ell)}(h^{i, \ell - 1})
\end{align}
where $e^i$ is the token embedding, $p^i$ is the positional encoding, and $h^{i,\ell}$ denotes the hidden state at the $\ell$-th Transformer layer. Then, the generator LLM $\mathcal{M}$ can be defined as:
\begin{align}
\mathcal{M}(s^{i+1} \mid s^{\leq i}) &= \text{softmax}(W h^{i,L}) \;,
\end{align}
where $h^{i,L}$ is the last layer's hidden state at position $i$, and $W$ is the parameter of the LLM head.

We extract the hidden state \(h^{i,\ell} \) from a predefined $l$-th Transformer layer at the last generated token, and denote it as \( h_n \) for the $n$-th generated sequence for simplicity.
This hidden state \( h_n \) serves as a representation of \( \mathbf{s}_n \), and we repeat this process across $N$ samplings of the generator LLM $\mathcal{M}$ as part of test-time computation scaling.

\paragraph{Constructing adjacency matrix}
Given a set of hidden states \( \{ h_1, h_2, \ldots, h_N \} \) for each generated sequence $\mathbf{s}_n$, we compute the pairwise cosine similarity between each pair of hidden states to measure semantic similarity for subsequent clustering. Specifically, the similarity between the \( m \)-th and \( n \)-th hidden states is defined as:
\begin{align}
a_{m,n} = \frac{h_m \cdot h_n}{\|h_m\| \, \|h_n\|}\;,
\label{eq:cosine-sim}
\end{align}
and $a_{m,n}$ construct the adjacency matrix \( A \in \mathbb{R}^{N \times N} \). 
We observe that the cosine similarity $a_{m,n}$ between hidden states can effectively replace NLI-based semantic similarity used in existing soft clustering methods~\cite{lin2024generating, nikitin2024kernel}, as shown in Section~\ref{sec:exp:analysis}.

\paragraph{Spectral clustering}

Given the resulting adjacency matrix \( A \in \mathbb{R}^{N \times N} \), we apply spectral clustering to identify semantic clusters among the generated sequences. 
First, we compute the symmetric normalized Laplacian as follows:
\begin{align}
L \coloneqq I - D^{-1/2} A D^{-1/2} \;,
\label{eq:laplacian}
\end{align}
where \( D \) is the diagonal degree matrix with entries \( D_{m,m} \coloneqq \sum_{n=1}^N a_{m,n} \). We then calculate the eigenvalues \( \lambda_1 \leq \lambda_2 \leq \cdots \leq \lambda_N \) and corresponding eigenvectors \( v_1, v_2, \ldots, v_N \) of \( L \). The number of clusters \( k \) is determined by counting the number of eigenvalues below a predefined threshold \( \tau \), i.e., \( k \coloneqq |\{ \lambda_i \mid \lambda_i < \tau \}| \), based on the following theoretical result:

\begin{table*}[tb!]
\centering
\begin{adjustbox}{width=1.0\textwidth}
\begin{tabular}{lccccccc}
\toprule 

\multicolumn{1}{c}{\multirow{2.5}{*}{Method}}  & \multicolumn{1}{c}{\multirow{2.5}{*}{\makecell{External \\ Model $\theta_{ext}$}}} & \multicolumn{2}{c}{BioASQ} & \multicolumn{2}{c}{SQuAD} & \multicolumn{2}{c}{TriviaQA} \\
\cmidrule{3-8}
& & AUROC $\uparrow$ &  AUARC $\uparrow$  & AUROC $\uparrow$ &  AUARC $\uparrow$  & AUROC $\uparrow$ &  AUARC $\uparrow$  \\ 
\midrule
P(True) \cite{kadavath2022language} & LLM  & 0.7565 &	0.6663 &	0.6744 &	0.3559 &	0.7781 &	\textbf{0.7926}  \\
PE \cite{malinin2020uncertainty} & - & 0.8102 &	0.7006 &	\underline{0.7413} &	\underline{0.3954} &	\underline{0.7839} &	\underline{0.7912}   \\
KLE \cite{nikitin2024kernel} & NLI &  0.8361 &	0.6960 &	0.7199 &	0.3891 &	0.7771 &	0.7755 \\
Deg \cite{lin2024generating} & NLI &  0.7740 &	0.6411 &	0.6780 &	0.3466 &	0.6343 &	0.7083  \\
EigV \cite{lin2024generating} & NLI & 0.5646 &	0.5198 &	0.5877 &	0.2943 &	0.6210 &	0.7010 \\
ECC \cite{lin2024generating} & NLI & 0.8238 &	0.7060 &	0.6783 &	0.3694 &	0.7150 &	0.7677  \\
NumSets \cite{kuhn2023semantic} & NLI & 0.8318 &	0.6794 &	0.7311 &	0.3751 &	0.7717 &	0.7700 \\
SE \cite{kuhn2023semantic} & NLI & 0.8232 &	0.6873 &	0.7148 &	0.3876 &	0.7554 &	0.7666  \\
DSE \cite{kuhn2023semantic} & NLI & 0.8338 &	0.6883 &	0.7288 &	0.3835 &	0.7783 &	0.7752  \\
\cellcolor[HTML]{EFEFEF}SE-LSC (Ours) &  \cellcolor[HTML]{EFEFEF}- &  \cellcolor[HTML]{EFEFEF}\underline{0.8602} & \cellcolor[HTML]{EFEFEF}\textbf{0.7222} & \cellcolor[HTML]{EFEFEF}0.7401 & \cellcolor[HTML]{EFEFEF}0.3941 & \cellcolor[HTML]{EFEFEF}0.7745 & \cellcolor[HTML]{EFEFEF}0.7867 \\  
\cellcolor[HTML]{EFEFEF}DSE-LSC (Ours) &  \cellcolor[HTML]{EFEFEF}- &  \cellcolor[HTML]{EFEFEF}\textbf{0.8654} & \cellcolor[HTML]{EFEFEF}\underline{0.7152} & \cellcolor[HTML]{EFEFEF}\textbf{0.7616} & \cellcolor[HTML]{EFEFEF}\textbf{0.3981} & \cellcolor[HTML]{EFEFEF}\textbf{0.7843} & \cellcolor[HTML]{EFEFEF}0.7855 \\

\bottomrule 
\end{tabular}
\end{adjustbox}
\caption{
Comparison of the effectiveness of uncertainty quantification methods in terms of AUROC and AUARC on three datasets using Llama3-8B-Instruct. \textbf{Bold} and \underline{underlined} values indicate the best and second-best performance in each setting, respectively.
    }
\label{tab:main-uq}
\end{table*}

\begin{theorem} \cite{von2007tutorial} The multiplicity of the eigenvalue 0 of \( L \) is equal to the number of connected components in the graph defined by the adjacency matrix \( A \).
\end{theorem}

In other words, when the adjacency matrix \( A \) is binary, the number of connected components---and thus the number of clusters---exactly equals the multiplicity of the zero eigenvalue. For continuous-valued \( A \), the graph typically forms a single connected component; however, the distribution of small eigenvalues of \( L \) can still be used to estimate the number of semantically distinct meanings among the generated sequences.

Lastly, we perform hard semantic clustering over generated sequences, facilitating their direct integration into existing methods for uncertainty quantification~\cite{kuhn2023semantic} and multi-step reasoning~\cite{lee2025semantic}. Specifically, we construct a spectral embedding using the first \( k \) eigenvectors \( v_1, \ldots, v_k \), and apply \( k \)-means clustering in this reduced space to assign each sequence  \( \mathbf{s}_n \) to a semantic cluster $c$. In contrast, soft clustering methods leverage elements in spectral clustering such as the eigenvalue distribution to quantify LLM uncertainty~\cite{lin2024generating}.

\section{Experiments}

In this section, we empirically evaluate whether the proposed LSC effectively leverages the outputs generated with scaled inference-time computation to improve performance and efficiency across various downstream LLM tasks.
To assess the robustness of LSC, we conduct experiments with multiple LLMs, including Llama3-8B-Instruct~\cite{grattafiori2024llama} and Mistral-7B-Instruct-v0.1~\cite{DBLP:journals/corr/abs-2310-06825}. 
Additional results are presented in Appendix~\ref{append:extend-results}.
In what follows, we present results on uncertainty quantification tasks (Section~\ref{sec:exp:uq}) and multi-step reasoning tasks (Section~\ref{sec:exp:reasoning}).

\subsection{Uncertainty Quantification}
\label{sec:exp:uq}
We evaluate the effectiveness of LSC in uncertainty quantification tasks that estimate LLM confidence by analyzing multiple outputs generated from a single input.

\paragraph{Baselines} 
We compare our method to a range of existing approaches for uncertainty quantification.
\sftype{P(True)}~\cite{kadavath2022language} estimates the probability of correctness by prompting additional query to LLM, while \sftype{PE}~\cite{malinin2020uncertainty} computes predictive entropy via Monte Carlo dropout without considering semantics between sequences.
Recent methods can be categorized into soft and hard semantic clustering approaches.
Soft clustering methods such as \sftype{KLE} \cite{nikitin2024kernel}, \sftype{Deg}, \sftype{EigV}, and \sftype{ECC} \cite{lin2024generating} perform spectral clustering using semantic similarity graphs constructed using external NLI models.
In contrast, hard clustering methods such as \sftype{NumSets}, \sftype{SE}, and \sftype{DSE}~\cite{kuhn2023semantic} group semantically equivalent sequences by identifying mutual entailment relations through external NLI models.
However, our methods, \sftype{SE-LSC} and \sftype{DSE-LSC}, replace the external NLI models in \sftype{SE} and \sftype{DSE} with latent hidden states from the LLM, enabling more efficient and effective uncertainty estimation. 
We further extend this substitution to soft clustering methods in Section~\ref{sec:exp:analysis}.

\paragraph{Datasets}
We evaluate our method on three question answering (QA) datasets, each representing a distinct domain. TriviaQA~\cite{joshi2017triviaqa} reflects the open-domain setting with diverse and noisy trivia questions collected from the web. 
SQuAD~\cite{rajpurkar2016squad} represents the general-domain setting, featuring clean and well-structured question–context pairs from Wikipedia. BioASQ~\cite{krithara2023bioasq} corresponds to the biomedical domain, consisting of factoid questions that require expert-level understanding. We follow the experimental setup used in \citet{nikitin2024kernel}.

\begin{table*}[t!]
\centering
\begin{adjustbox}{width=1.0\textwidth}
\begin{tabular}{clccccccc}
\toprule 
\multicolumn{1}{c}{\multirow{3}{*}{Benchmark}} & \multicolumn{1}{c}{\multirow{3}{*}{Method}}& \multicolumn{1}{c}{\multirow{3.5}{*}{\makecell{External \\ Model \\ $\theta_{ext}$}}} & \multicolumn{3}{c}{Llama3-8B-Instruct} & \multicolumn{3}{c}{Mistral-7B-Instruct} \\
\cmidrule{4-9}
& & & Accuracy $\uparrow$ &  \makecell[c]{\# of LLM \\ inferences $\downarrow$}  & \makecell[c]{\# of NLI \\ inferences $\downarrow$}  & Accuracy $\uparrow$ & \makecell[c]{\# of LLM \\ inferences $\downarrow$}  & \makecell[c]{\# of NLI \\ inferences $\downarrow$}  \\ 
\midrule
\multirow{4}{*}{GSM8K}& ToT \cite{yao2023tree}&-  & 0.79	& 104.80	&- &0.61 	& 122.16 &-	  \\
& RAP \cite{hao2023reasoning}&- & 0.83	& 128.40	&- &\underline{0.67}	& 161.86 &-	  \\
& SExp \cite{lee2025semantic}&NLI  &\underline{0.85}	&\underline{82.63}	&62.86 &\textbf{0.68}	&\underline{98.79} &78.26	  \\
& \cellcolor[HTML]{EFEFEF}SExp-LSC (Ours)& \cellcolor[HTML]{EFEFEF}-	& \cellcolor[HTML]{EFEFEF}\textbf{0.86}	& \cellcolor[HTML]{EFEFEF}\textbf{81.40}	& \cellcolor[HTML]{EFEFEF}-	& \cellcolor[HTML]{EFEFEF}\underline{0.67}		&\cellcolor[HTML]{EFEFEF}\textbf{84.38} 	& \cellcolor[HTML]{EFEFEF}-	 \\
\midrule
\multirow{4}{*}{ARC}& ToT&-  & 0.80	& 149.59	&- &	 0.62 	& 221.36 &-	  \\
& RAP&- & 0.81	& 196.96	&- &\textbf{0.71} 	&  313.20 &-	  \\
& SExp &NLI  &\textbf{0.83}	&\underline{96.32}	&70.42 &\textbf{0.71} 	&\underline{155.13} &121.67	  \\
& \cellcolor[HTML]{EFEFEF}SExp-LSC (Ours)& \cellcolor[HTML]{EFEFEF}-	& \cellcolor[HTML]{EFEFEF}\textbf{0.83}	& \cellcolor[HTML]{EFEFEF}\textbf{82.97}	& \cellcolor[HTML]{EFEFEF}-	& \cellcolor[HTML]{EFEFEF}\textbf{0.71}		&  \cellcolor[HTML]{EFEFEF}\textbf{120.87}	& \cellcolor[HTML]{EFEFEF}-	 \\
\bottomrule
\end{tabular}
\end{adjustbox}
\caption{
Comparison of accuracy and inference efficiency for reasoning methods on both GSM8K and ARC datasets using Llama3-8B-Instruct and Mistral-7B-Instruct. \textbf{Bold} values denote the best results, and \underline{underlined} values indicate the second-best.
}
\label{tab:main-reason}
\end{table*}


\begin{table*}[t!]
\centering
\label{tab:main}
\begin{adjustbox}{width=1.0\textwidth}
\begin{tabular}{ccccccccc}
\toprule

Clustering  & Model Type & Model Name  & Context & Memory (GB) $\downarrow$ & Latency (sec) $\downarrow$ & F1 Score $\uparrow$ & Precision $\uparrow$  &  Recall $\uparrow$  \\ 
\midrule
\multirow{2}{*}{BDEC}& \multirow{2}{*}{NLI} & \multirow{2}{*}{DeBERTa-MNLI-Large}  & \xmark  & \multirow{2}{*}{1.5212} & 0.1298 & 0.4365 & 0.9558 & 0.3865 \\
& & & \cmark &  & 0.1447 & 0.6904 & 0.6232 & 0.9442 \\
\cmidrule(lr){1-9}
\multirow{13}{*}{\makecell{Spectral\\Clustering}} & \multirow{7}{*}{Sentence BERT}  & \multirow{2}{*}{all-MiniLM-L6-v2} & \xmark & \multirow{2}{*}{0.0926} & 0.0225 & 0.7058 & 0.7414 & 0.8070 \\
&  &  & \cmark & & 0.0217 & 0.6888 & 0.6603 & 0.8816 \\
\cmidrule(lr){3-9}
 &  & \multirow{2}{*}{all-mpnet-base-v2} & \xmark & \multirow{2}{*}{0.4618} & 0.0372 & 0.7198 & 0.7043 & 0.8703 \\
&  & & \cmark &  & 0.0361 & 0.7078 & 0.6653 & 0.9101 \\
\cmidrule(lr){3-9}
 &  & \multirow{2}{*}{gtr-t5-xxl} & \xmark & \multirow{2}{*}{18.1341} & 0.1358 & 0.7333 & 0.7407 & 0.8468 \\
&  & & \cmark &  & 0.3287 & 0.7150 & 0.6497 & 0.9353 \\
\cmidrule(lr){2-9}
 & \multirow{6}{*}{Decoder-only LLM} & \multirow{2}{*}{LLaMa3.2-1B} & \xmark & \multirow{2}{*}{4.6127} & 0.0497 & 0.7098 & 0.7234 & 0.8246 \\
&  & & \cmark &  & 0.0820 & 0.8160 & 0.8369 & 0.8805 \\
\cmidrule(lr){3-9}
& & \multirow{2}{*}{Mistral-7B} & \xmark & \multirow{2}{*}{19.9969} & 4.4361 & 0.6998 & 0.6812 & 0.8774 \\
&  & & \cmark &  & 4.9110 & 0.8415 & 0.8816 & 0.8844 \\
\cmidrule(lr){3-9}
& & \cellcolor[HTML]{EFEFEF}LSC & \cellcolor[HTML]{EFEFEF}\cmark & \cellcolor[HTML]{EFEFEF}$\boldsymbol{\approx 0}$ & \cellcolor[HTML]{EFEFEF}\textbf{0.0069} & \cellcolor[HTML]{EFEFEF}\textbf{0.8789} & \cellcolor[HTML]{EFEFEF}0.8825 & \cellcolor[HTML]{EFEFEF}0.9367 \\

\bottomrule 
\end{tabular}
\end{adjustbox}
\caption{
Comparison of clustering performance for NLI, sentence-embedding, and LLM-based models in multi-step reasoning. We present F1 score, precision, recall, latency (sec), and memory usage (GB) to evaluate both clustering quality and computational efficiency. \textbf{Bold} values indicate the best performance in each metric.
}
\label{tab:clustering-perform}
\end{table*}

\paragraph{Results}
Table~\ref{tab:main-uq} compares various uncertainty quantification methods using the Area Under the Receiver Operating Characteristic (AUROC) and the Area Under the Accuracy-Rejection Curve (AUARC). 
AUROC evaluates how accurately a method ranks predictions by computed uncertainty, assigning higher uncertainty to incorrect answers.
In contrast, AUARC reflects calibration by tracking how accuracy improves as high-uncertainty predictions are progressively removed based on a threshold.
While existing NLI-based baselines demonstrate moderate performance, our proposed methods, \sftype{SE-LSC} and \sftype{DSE-LSC}, consistently achieve the highest AUROC scores across all settings and yield the best or comparable AUARC values. 
Notably, our methods outperform or match existing baselines without requiring additional computational costs such as NLI models. These results suggest our method provides more reliable uncertainty estimation in uncertainty quantification tasks. Additional results with more LLMs are reported in Appendix~\ref{append:extend-results}.

\subsection{LLM Reasoning} \label{sec:exp:reasoning}
To evaluate the effectiveness and efficiency of LSC in enhancing reasoning capabilities, we investigate multi-step reasoning tasks that use tree search as a test-time computation scaling strategy, where LSC is applied to merge semantically redundant reasoning paths.
Specifically, tree search-based reasoning approaches explore multiple intermediate reasoning paths using algorithms such as beam search and Monte Carlo tree search (MCTS).

\paragraph{Baselines} 
We consider three baseline methods for tree search-based reasoning.
Tree-of-Thoughts~(\sftype{ToT})~\cite{yao2023tree} and Reasoning-via-Planning~(\sftype{RAP})~\cite{hao2023reasoning} are well-known approaches that leverage beam search and MCTS, respectively.
Semantic Exploration~(\sftype{SExp})~\cite{lee2025semantic} is a more recent work that improves tree search efficiency by adopting semantic clustering to avoid semantically redundant exploration on reasoning trees.
We note that \sftype{SExp} utilizes the BDEC algorithm~\cite{kuhn2023semantic} with an external NLI model, DeBERTa-MNLI-Large~\cite{he2020deberta}, for semantic clustering.

\paragraph{Datasets}
To evaluate the reasoning capabilities of LLMs, we use two benchmarks: GSM8K~\cite{cobbe2021gsm8k}, a dataset consisting of 8.5k math word problems that require multi-step mathematical reasoning, and the AI2 Reasoning Challenge (ARC), which contains 7.8k science questions derived from grade-school science exames.
Following the evaluation setup of \sftype{SExp}~\cite{lee2025semantic}, we randomly sample 400 examples from the test sets of GSM8K and ARC for evaluation.

\paragraph{Results}
In Table~\ref{tab:main-reason}, we evaluate the effect of LSC on multi-step reasoning tasks.
\sftype{ToT} and \sftype{RAP} require a large number of LLM inferences, as they expand and explore semantically duplicate reasoning paths.
\sftype{SExp} reduces this computational cost by merging semantically redundant paths using an external NLI model, but still relies on external supervision. 
In contrast, our \sftype{SExp-LSC} achieves comparable accuracy without any external model by leveraging internal LLM representations. 
Notably, on the ARC dataset, \sftype{SExp-LSC} reduces the number of LLM inferences by 22.58\% while maintaining accuracy, demonstrating the efficiency and effectiveness of our approach.
This reduction in LLM inferences stems from LSC’s strong clustering performance, enabled by its use of context-aware semantic representations to merge semantically identical reasoning paths.

\begin{figure*}[ht!]
     \centering
    \includegraphics[width=\textwidth]{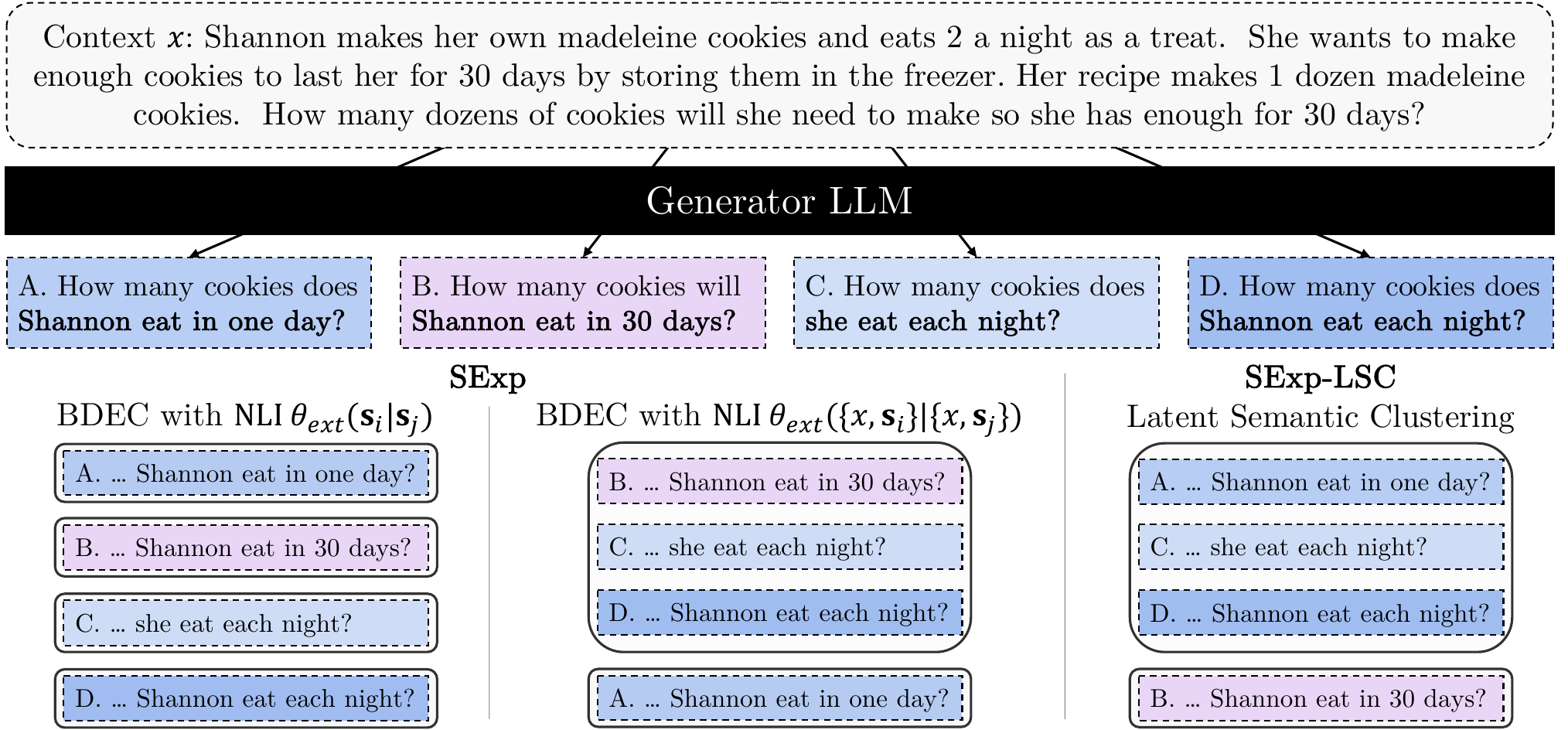}
    \vspace{-1em}
    \caption{Example of semantic clustering from GSM8K using Llama3-8B-Instruct  comparing \sftype{SExp} and \sftype{SExp-LSC}. The NLI-based model fails to group context-dependent semantically identical sequences (e.g., A, C and D), while LSC, using context-conditioned hidden representations, correctly clusters semantically equivalent sequences under a given context. This highlights a key limitation of NLI-based methods: without context, they miss context-dependent meaning, and with context, they often fail to properly process it along with the generated sequences. The full constructed reasoning trees are shown in Figure~\ref{fig:constructed-tree}.}\label{fig:se-case}
    \vspace{-3mm}
\end{figure*}

\section{Further Analyses}\label{sec:exp:analysis}
\paragraph{Effect of LSC on clustering}
To evaluate the effectiveness and efficiency of leveraging hidden states from the generator LLM for semantic clustering, we compare LSC with existing NLI- and embedding-based approaches.
Specifically, we evaluate clustering performance on the first-level sequences of reasoning trees generated by Llama3-8B-Instruct on GSM8K. To construct clustering labels, we use Llama3-70B-Instruct to generate pairwise semantic equivalence labels by using a prompt, as illustrated in Figure~\ref{fig:appendix-prompt-label-gen}. Clustering performance is measured by computing precision, recall, and F1 score for the generated sequences of each example and averaging across all cases. We also assess computational efficiency by measuring memory usage and latency. To evaluate how well each model handles contextual information, we conduct experiments both with and without including the context as part of the input.

\begin{figure}[t!]
    \centering
    \begin{subfigure}[b]{0.49\columnwidth}
        \centering
        \includegraphics[width=\linewidth]{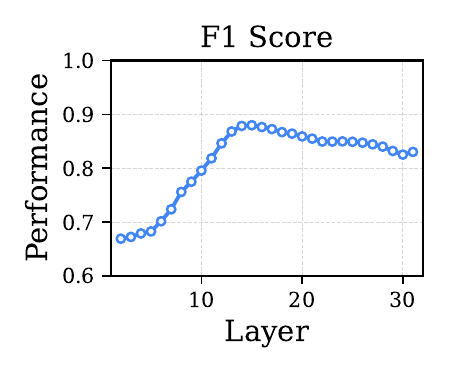}
        \label{fig:layer-by-layer:f1}
    \end{subfigure}
    \hfill
    \begin{subfigure}[b]{0.49\columnwidth}
        \centering
        \includegraphics[width=\linewidth]{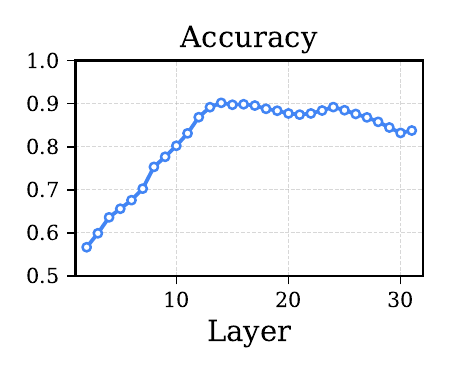}
        \label{fig:layer-by-layer:acc}
    \end{subfigure}
    \vspace{-13mm}
    \caption{Layer-wise clustering performance using LLaMA3-8B-Instruct hidden states. Intermediate and later layers demonstrate higher F1 scores and accuracy than earlier layers.}
    \label{fig:layer-by-layer}
\end{figure}

NLI-based models show a trade-off between precision and recall depending on whether context is included, while BERT-based embeddings struggle to capture context-dependent semantics, resulting in suboptimal clustering as demostrated in Table~\ref{tab:clustering-perform}.
In contrast, LLM-based embeddings benefit significantly from context: both precision and recall increase when contextual information is included.
Our LSC method, which leverages internal LLM representations, achieves the highest F1 score with negligible memory and latency overhead, maintaining semantic fidelity while offering superior efficiency. 
As shown in prior work~\cite{skean2025layer}, we also observe that intermediate and later layers in LLMs provide robust and effective semantic representations as shown in Figure~\ref{fig:layer-by-layer}.

\begin{figure}[t!]
    \centering
    \includegraphics[width=.94\columnwidth]{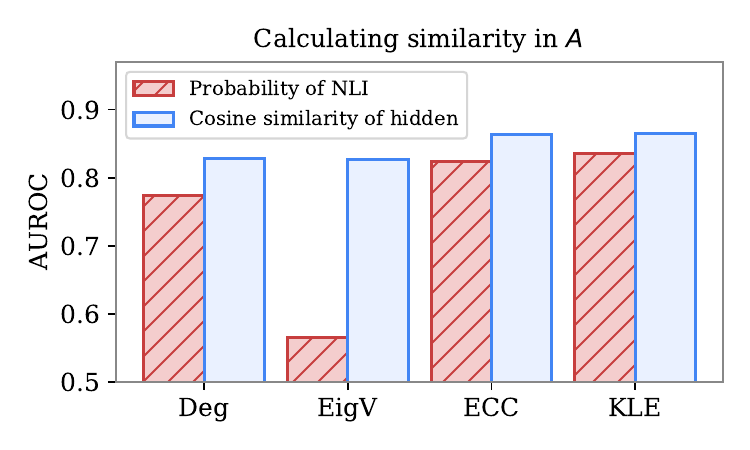}
    \vspace{-5mm}
    \caption{Comparison of different adjacency matrix construction methods for soft clustering-based uncertainty quantification approaches leveraging semantic similarities between generated sequences.}
    \label{fig:soft-edge}
\end{figure}

\paragraph{Context-aware semantic clustering}
To qualitatively assess whether LSC captures context-aware semantics, Figure~\ref{fig:se-case} presents an example from GSM8K using Llama3-8B-Instruct. The figure illustrates the first level of reasoning trees generated by \sftype{SE} and \sftype{SE-LSC}. When the shared context $x$ is excluded from the input, the NLI-based clustering model in \sftype{SE} lacks context awareness and instead relies solely on surface-level sentence semantics, resulting in all sequences being grouped into separate clusters. Indeed, in the absence of shared context, sequences A, C, and D appear to convey distinct meanings, despite being semantically equivalent when the context is considered. However, even when the context is included, the model still fails to accurately identify and cluster semantically equivalent sentences, as it does not effectively incorporate contextual information. In contrast, LSC successfully performs context-aware semantic clustering by leveraging the hidden representations from the generator LLM, which are inherently conditioned on the input context.

\paragraph{Semantic similarity as soft edge}

Subsequent works of~\citet{kuhn2023semantic} consider semantic similarity, rather than strict equivalence, for uncertainty quantification. \sftype{KLE}~\cite{nikitin2024kernel}, \sftype{Deg}, \sftype{EigV}, and \sftype{ECC}~\cite{lin2024generating} leverage spectral clustering based on an adjacency matrix constructed using an NLI model, enabling uncertainty estimation without hard semantic clustering. Therefore, their adjacency matrix can be substituted with our adjacency matrix in Equation~\eqref{eq:cosine-sim}, which is based on cosine similarity between hidden states. We evaluate these methods to compare adjacency matrix construction based on NLI with our approach on the BioASQ dataset using Llama3-8B-Instruct. As shown in Figure~\ref{fig:soft-edge}, our method described in Equation~\eqref{eq:cosine-sim} achieves superior performance in both AUROC and AUARC, suggesting the potential of computing semantic similarity from hidden states of the generator LLMs.

\begin{figure}[t]
    \centering

    \begin{subfigure}[b]{0.48\columnwidth}
        \includegraphics[width=\linewidth]{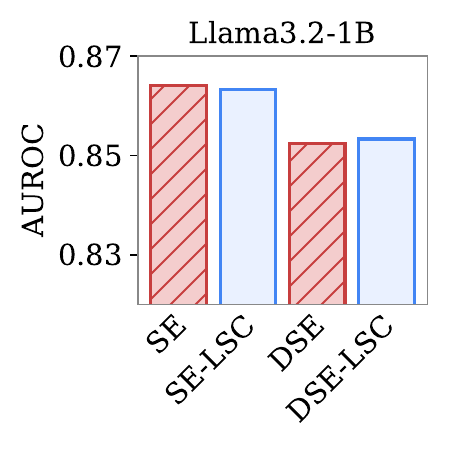}
    \end{subfigure}
    \hfill
    \begin{subfigure}[b]{0.48\columnwidth}
        \includegraphics[width=\linewidth]{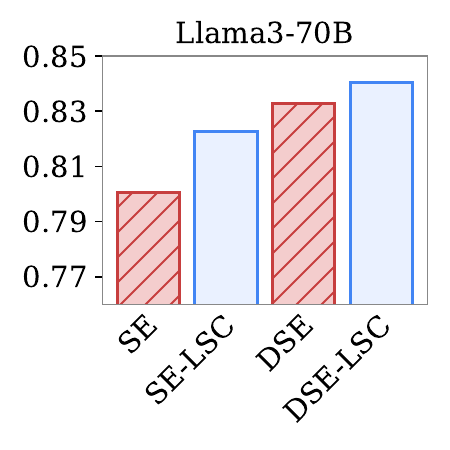}
    \end{subfigure}
    \vspace{-4mm}

    \caption{Effect of LLM size on the effectiveness of LSC. Comparison of AUROC scores for SE, DSE, SE-LSC, and DSE-LSC on the BioASQ dataset using Llama3.2-1B-Instruct and Llama3-70B-Instruct.}
    \label{fig:smaller-and-larger}
\end{figure}

\paragraph{Smaller and larger LLMs}
To examine the robustness of LSC with respect to LLM size, we conduct experiments using Llama3.2-1B-Instruct~\cite{meta2024llama32} and Llama3-70B-Instruct~\cite{grattafiori2024llama} on the BioASQ dataset, performing uncertainty quantification. For the smaller LLM (\textit{i.e.,} Llama3.2-1B-Instruct), \sftype{SE-LSC} and \sftype{DSE-LSC} shows comparable performance in AUROC and AUARC to the \sftype{SE} and \sftype{DSE} as shown in Figure~\ref{fig:smaller-and-larger}. In contrast, for the larger LLM (\textit{i.e.,} Llama3-70B-Instruct), \sftype{SE-LSC} and \sftype{DSE-LSC} shows greater improvements over the baselines, SE and DSE. These results suggest that LSC may be more effective with larger LLMs.

\begin{table}[t!]
\centering
\begin{adjustbox}{width=\linewidth}
\begin{tabular}{lcccc}
\toprule
\makecell{Clustering method} & Precision $\uparrow$ & Recall $\uparrow$ & F1 Score $\uparrow$ \\

\midrule

$k$-means & 0.8865 &	0.7232 &	0.7387\\
AHC & 0.9308 &	0.8837 &	0.8664\\
DBSCAN & 0.8383 &	0.9660 &	0.8676\\
\cellcolor[HTML]{EFEFEF}LSC (Ours) & 
\cellcolor[HTML]{EFEFEF}0.8825 & 
\cellcolor[HTML]{EFEFEF}0.9367 & 
\cellcolor[HTML]{EFEFEF}\textbf{0.8789} \\

\bottomrule
\end{tabular}
\end{adjustbox}
\caption{Comparison of clustering methods, $k$-means, AHC, DBSCAN, and LSC, using LLaMA3-8B-Instruct hidden states for semantic clustering on the GSM8K dataset. We report precision, recall, and F1 score. \textbf{Bold} value indicates the best performance.} \label{tab:diff-clustering-methods}
\end{table}

\paragraph{Different clustering algorithms}
We compare LSC with standard clustering algorithms—$k$-means, Agglomerative Hierarchical Clustering (AHC), and DBSCAN—using the same hidden states from Llama3-8B-Instruct. For all baselines, we report the best F1 score obtained by hyperparameter search, except for $k$-means, where the number of clusters is determined using the elbow method. As shown in Table~\ref{tab:diff-clustering-methods}, LSC achieves the highest F1 score of 0.8789, outperforming other clustering methods and indicating its practical effectiveness for semantic clustering in downstream tasks.

\section{Conclusion}

In this work, we propose Latent Semantic Clustering (LSC), a lightweight and context-aware approach for identifying semantically equivalent outputs generated by LLMs for test-time computation scaling. 
Rather than relying on external models as in prior works, LSC leverages the generator LLM’s internal hidden states to perform context-aware semantic clustering with minimal overhead.
Our experiments show that LSC achieves strong performance on key tasks like uncertainty quantification and multi-step reasoning, while substantially reducing computational cost.
LSC consistently outperforms prior semantic clustering methods in both clustering quality and computational efficiency. 
We believe LSC offers a practical and scalable solution for enhancing the reliability and quality of LLM responses, thereby improving test-time computation scaling strategies.

\section*{Limitations}

While our approach demonstrates strong performance and efficiency, there are limitations that suggest directions for future improvement. First, our method relies on the hidden state from a single intermediate layer and the last token, which provides an efficient and effective representation in practice. However, incorporating multiple layers, aggregating token-level information, or utilizing attention patterns may further enhance representation quality and is a promising direction for future work. Second, LSC requires access to the generator model’s hidden states, and thus is currently applicable only to white-box LLMs. While this design enables improvements in efficiency and effectiveness, it may limit applicability in scenarios involving black-box models. Exploring alternatives that approximate internal representations without direct access could help broaden the scope of LSC.



\bibliography{custom}

\clearpage

\appendix

\section*{Appendix}\label{sec:appendix}

\renewcommand{\thefigure}{A\arabic{figure}}
\setcounter{figure}{0}

\renewcommand{\thetable}{A\arabic{table}}
\setcounter{table}{0}

\begin{figure*}[t!]
     \centering
    \includegraphics[width=\textwidth]{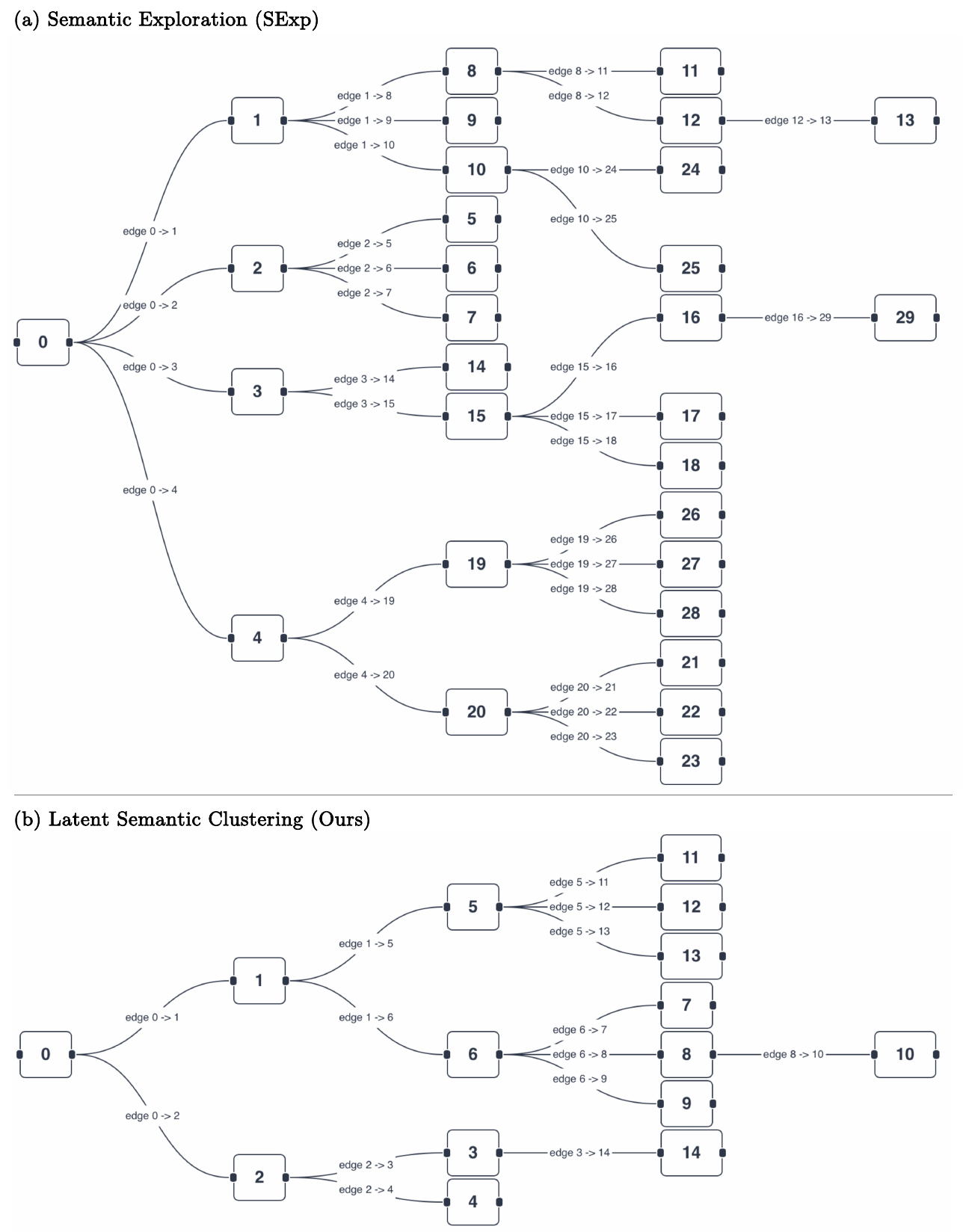}
    \vspace{-1em}
    \caption{Reasoning trees constructed from GSM8K using Llama3-8B-Instruct, comparing \sftype{SExp} and \sftype{SExp-LSC}. \sftype{SExp-LSC} constructs shallower trees by leveraging context-dependent semantics, avoiding semantically redundant paths. A concrete example of generated sequences is shown in Figure~\ref{fig:se-case}.}\label{fig:constructed-tree}
    \vspace{-3mm}
\end{figure*}

\section{Related Work}\label{append:related-work}
\paragraph{Scaling LLM test-time computation}
Leveraging increased computation at inference time-through parallel or sequential sampling-allows LLMs to enhance the quality and reliability of their outputs on various NLP tasks~\cite{wei2022chain, yao2023tree, snell2025scaling}. A simple yet powerful approach generates multiple outputs for a given prompt and identifies the most consistent one among the outputs~\cite{wang2022self, xiong2024can}. Recently, several studies have proposed incorporating semantic relationship to identify significant outputs in open-ended natural language generation tasks by considering semantics. For example, \citet{kuhn2023semantic} generates multiple sequences and clusters them to estimate LLM uncertainty. Similarly, studies on LLM reasoning identify semantically equivalent sequences among multiple samples to avoid redundant reasoning paths and encourage exploration of more significant ones~\cite{lee2025semantic, wang2025don}.

\paragraph{Semantic clustering in LLMs.}
Existing methods for semantic clustering typically have relied on external models to evaluate semantic relationship between textual sequences. For example, \citet{kuhn2023semantic} employ a natural language inference (NLI) model to identify these relationships and performs clustering based on the model's outputs. Subsequent works also leverage the NLI models and adopt spectral clustering to quantify uncertainty~\cite{lin2024generating, nikitin2024kernel, qiu2024semantic}. Alternatively, \citet{abdaljalil2025sindex} utilize Sentence-BERT-based embedding model and perform clustering based on cosine similarity between the embeddings. However, these approaches rely on external models and incur additional computational overhead, which can hinder scalability and efficiency. In addition, the external models such as NLI and embedding models often struggle to effectively handle model inputs consisting of context and generated sequences. Notably, prior studies focus solely on uncertainty quantification, overlooking the extension to different LLM tasks.

\paragraph{Leveraging hidden states in LLMs}
Recent studies have explored the utility of LLM hidden states across various downstream tasks.
In text clustering, embeddings derived from LLMs exhibit rich semantic information, outperforming conventional sentence embeddings in capturing contextual similarity~\cite{petukhova2025text}.
However, this work focuses on clustering a given set of input texts, rather than analyzing internal representations across generated outputs.
Other lines of work use hidden states to estimate uncertainty for hallucination detection~\cite{kossen2024semantic,hansemantic}, but often rely on training additional lightweight multi-layer perceptrons on top of the hidden states, which introduces computational overhead.
More recently, several studies have shown that LLMs can reason and self-evaluate directly in the latent space, without producing explicit outputs~\cite{hao2024training, wang2025latent}. 
However, we instead leverage latent hidden states for clustering generated outputs, enabling applications to test-time scaling tasks without additional training or supervision.

\section{Experimental Settings}\label{append:extend-settings}
The hyper-parameters used in our experiments are listed in Table~\ref{appendix:tab:hyperparameter-uq} for the uncertainty quantification tasks. The clustering threshold $\tau$ for LSC is selected based on performance on a validation set of 100 examples per task. Table~\ref{appendix:tab:hyperparameter-reasoning} lists the hyper-parameters used for multi-step reasoning tasks.

\section{Extended Experimental Results}\label{append:extend-results}
Table~\ref{appendix:tab:main_results_UQ} presents extended experimental results for the uncertainty quantification task, evaluated using three datasets across two models: Llama3-8B-Instruct and Mistral-7B-Instruct.

\section{Prompt}\label{append:prompt}
Figure \ref{fig:appendix-prompt-uq} illustrates the 5-shot in-context prompt used for uncertainty quantification.
Figures \ref{fig:appendix-prompt-gsm} and \ref{fig:appendix-prompt-arc} show the 1-shot in-context prompts used for semantic exploration. These are identical to the prompts used in \sftype{SExp}~\cite{lee2025semantic} experiments.

\begin{table*}[htp!]
    \centering
    \resizebox{0.6\linewidth}{!}{%
    \begin{tabular}{ccc}
        \toprule
        Setting & Hyperparameter & Value \\
        \midrule
        \multirow{3}{*}{\makecell{LLM-related \\ setting}} & temperature & $1.0$ \\
         & top-$k$  & $50$ \\
         & top-$p$ & $1.0$ \\
        \midrule
        \multirow{2}{*}{General setting} & batch size & $1$ \\
         & the number of examples (prompts) & $1$ \\
        \midrule
        \multirow{1}{*}{\makecell{Deg, EigV, ECC}}
         & threshold for eigenvalues & * \\
        \midrule
        Ours (LSC) & threshold for clustering & * \\ 
        \bottomrule
    \end{tabular}
    }
    \caption{The default hyperparameters used in our experiments are categorized into LLM-related, general, and method-specific settings. * indicates that the hyperparameter is selected based on the best threshold determined from the validation set.}
    \label{appendix:tab:hyperparameter-uq}
\end{table*}

\begin{table*}[htp!]
    \centering
    \begin{tabular}{ccc}
        \toprule
        Setting & Parameter & Value \\
        \midrule
         \multirow{3}{*}{LLM-related setting} &  temperature & $0.8$ \\
         & top-$k$  & $50$ \\
         & top-$p$ & $0.95$ \\
         \midrule
         \multirow{2}{*}{General setting } & batch size & $1$ \\
         & the number of examples (prompts) & $1$ \\
         \midrule
         \multirow{6}{*}{MCTS setting} & depth limit & $5$ \\
         & the number of iterations & $10$ \\
         & the number of actions & $4$ \\
         & reward alpha & $0.5$ \\
         & the number of confidences & $8$ \\
         & a default value of reward confidence  & $0.8$ \\
         \midrule
         \multirow{2}{*}{ToT setting} & beam size & 3 \\
         & depth limit & $5$ \\
         \bottomrule
    \end{tabular}
    \caption{Default hyper-parameters for SEXp, RAP, ToT. Parameters are grouped into LLM-related, general, and method-specific settings, following the settings used in~\sftype{SExp}~\cite{lee2025semantic}}
    \label{appendix:tab:hyperparameter-reasoning}
\end{table*}

\begin{table*}[t]
\centering
\small
\begin{tabular}{lclcccccc}
\toprule
\multirow{2}{*}{Method} & \multirow{2}{*}{Module} & \multicolumn{2}{c}{BioASQ} & \multicolumn{2}{c}{SQuAD} & \multicolumn{2}{c}{TriviaQA} \\
\cmidrule(lr){3-4} \cmidrule(lr){5-6} \cmidrule(lr){7-8}
& & AUROC ↑ & AUARC ↑ & AUROC ↑ & AUARC ↑ & AUROC ↑ & AUARC ↑ \\
\midrule
\multicolumn{8}{c}{Llama3-8B-Instruct} \\
\midrule
P(True) \cite{kadavath2022language} & LLM & 0.7565 & 0.6663 & 0.6744 & 0.3559 & 0.7781 & \textbf{0.7926} \\
PE \cite{malinin2020uncertainty} & - & 0.8102 & 0.7006 & \underline{0.7413} & \underline{0.3954} & \underline{0.7839} & \underline{0.7912} \\
KLE \cite{nikitin2024kernel} & NLI & 0.8361 & 0.6960 & 0.7199 & 0.3891 & 0.7771 & 0.7755 \\
Deg \cite{lin2024generating} & NLI & 0.7740 & 0.6411 & 0.6780 & 0.3466 & 0.6343 & 0.7083 \\
EigV \cite{lin2024generating} & NLI & 0.5646 & 0.5198 & 0.5877 & 0.2943 & 0.6210 & 0.7010 \\
ECC \cite{lin2024generating} & NLI & 0.8238 & 0.7060 & 0.6783 & 0.3694 & 0.7150 & 0.7677 \\
NumSets \cite{kuhn2023semantic} & NLI & 0.8318 & 0.6794 & 0.7311 & 0.3751 & 0.7717 & 0.7700 \\
SE \cite{kuhn2023semantic} & NLI & 0.8232 & 0.6873 & 0.7148 & 0.3876 & 0.7554 & 0.7666 \\
DSE \cite{kuhn2023semantic} & NLI & 0.8338 & 0.6883 & 0.7288 & 0.3835 & 0.7783 & 0.7752 \\
\cellcolor[HTML]{EFEFEF}SE-LSC (Ours) & \cellcolor[HTML]{EFEFEF}- & \cellcolor[HTML]{EFEFEF}\underline{0.8602} & \cellcolor[HTML]{EFEFEF}\textbf{0.7222} & \cellcolor[HTML]{EFEFEF}0.7401 & \cellcolor[HTML]{EFEFEF}0.3941 & \cellcolor[HTML]{EFEFEF}0.7745 & \cellcolor[HTML]{EFEFEF}0.7867 \\
\cellcolor[HTML]{EFEFEF}DSE-LSC (Ours) & \cellcolor[HTML]{EFEFEF}- & \cellcolor[HTML]{EFEFEF}\textbf{0.8654} & \cellcolor[HTML]{EFEFEF}\underline{0.7152} & \cellcolor[HTML]{EFEFEF}\textbf{0.7616} & \cellcolor[HTML]{EFEFEF}\textbf{0.3981} & \cellcolor[HTML]{EFEFEF}\textbf{0.7843} & \cellcolor[HTML]{EFEFEF}0.7855 \\
\midrule
\multicolumn{8}{c}{Mistral-7B-Instruct} \\
\midrule
P(True) & LLM & 0.7217 & 0.5113 & 0.6718 & 0.3006 & 0.7905 & 0.6800 \\
PE & - & 0.5233 & 0.3831 & 0.7094 & 0.3149 & 0.7653 & 0.6555 \\
KLE & NLI & 0.8575 & 0.5745 & 0.7536 & \textbf{0.3317} & \underline{0.8310} & \underline{0.7025} \\
Deg & NLI & 0.7927 & 0.5402 & 0.6296 & 0.2675 & 0.6434 & 0.5818 \\
EigV & NLI & 0.5313 & 0.3704 & 0.4838 & 0.1972 & 0.5388 & 0.5096 \\
ECC & NLI & 0.7508 & 0.5268 & 0.5911 & 0.2628 & 0.6491 & 0.6047 \\
NumSets & NLI & 0.8553 & 0.5503 & 0.7519 & 0.3090 & 0.8164 & 0.6778 \\
SE & NLI & 0.8571 & \underline{0.5759} & 0.7516 & \underline{0.3313} & \textbf{0.8338} & \textbf{0.7081} \\
DSE & NLI & 0.8569 & 0.5587 & 0.7517 & 0.3188 & 0.8245 & 0.6927 \\
\cellcolor[HTML]{EFEFEF}SE-LSC (Ours) & \cellcolor[HTML]{EFEFEF}- & \cellcolor[HTML]{EFEFEF}\underline{0.8605} & \cellcolor[HTML]{EFEFEF}\textbf{0.5866} & \cellcolor[HTML]{EFEFEF}\underline{0.7538} & \cellcolor[HTML]{EFEFEF}0.3284 & \cellcolor[HTML]{EFEFEF}0.8234 & \cellcolor[HTML]{EFEFEF}0.6964 \\
\cellcolor[HTML]{EFEFEF}DSE-LSC (Ours) & \cellcolor[HTML]{EFEFEF}- & \cellcolor[HTML]{EFEFEF}\textbf{0.8696} & \cellcolor[HTML]{EFEFEF}0.5565 & \cellcolor[HTML]{EFEFEF}\textbf{0.7584} & \cellcolor[HTML]{EFEFEF}0.3267 & \cellcolor[HTML]{EFEFEF}0.8161 & \cellcolor[HTML]{EFEFEF}0.6870 \\
\bottomrule
\end{tabular}
\caption{Comparison of uncertainty quantification methods based on AUROC across three datasets using two LLMs: LLaMa3-8B and Mistral-7B. \textbf{Bold} and \underline{underlined} values indicate the best and second-best performance in each setting, respectively.}
\label{appendix:tab:main_results_UQ}
\end{table*}

\definecolor{customgrayblue}{HTML}{ECEFF6}
\definecolor{customframeblue}{HTML}{6795ED}

\newtcolorbox{prompt}[1]{
  colback=customgrayblue,
  colframe=customframeblue,
  fonttitle=\bfseries,
  title={#1},
  left=2pt,
  right=2pt,
  fontupper=\small,
}

\newcommand{\promptEmphExample}{\it \color{blue!70!white}}

\begin{figure*}[hp!]
\begin{prompt}{An answer generation prompt for BioASQ}
Input data is Answer the following question as briefly as possible.

{
\promptEmphExample

Question: Is there a sequence bias in MNase digestion patterns?

Answer: yes

Question: Is the transcriptional regulator BACH1 an activator or a repressor?

Answer: Repressor

Question: Which fusion protein is involved in the development of Ewing sarcoma?

Answer: EWS/FLI1

Question: Is PLK2 involved in alpha-synuclein phosphorylation in the nervous system?

Answer: yes

Question: What gene is mutated in Familial Mediterranean Fever?

Answer: MEFV gene
}

Question: What are 'vildagliptin', 'sitagliptin', 'saxagliptin', 'alogliptin', 'linagliptin', and 'dutogliptin'?

Answer:

\end{prompt}

\begin{prompt}{An answer generation prompt for SQuAD}

Input data is Answer the following question as briefly as possible.

{ \promptEmphExample

Question: Which side of the road do vehicles on Cyprus drive on?

Answer: left-hand

Question: What did John Rawls publish?

Answer: A Theory of Justice

Question: What is the attempt to understand other societies on their own terms?

Answer: cultural relativism

Question: Which park does 27th Street pass through between Ninth and Tenth Avenues?

Answer: Chelsea

Question: What caused the constant linear velocity?

Answer: Noel Pemberton Billing's patented add-on governor device

}

Question: Following the end of the second World War, what was a still a popular theme among films makers in Burma ?

Answer:
\end{prompt}
\begin{prompt}{A reward generation prompt for TriviaQA}

Input data is Answer the following question as briefly as possible.

{\promptEmphExample
Question: Who founded the Jaguar motor company?

Answer: william lyons

Question: The name of which Russian spacecraft means 'travelling companion' or 'satellite'?

Answer: sputnik

Question: Which record label recorded The Supremes and The Jackson 5?

Answer: motown

Question: What was the Russian City of Nizhny Novgorod called between 1932 and 1990?

Answer: gorky

Question: "Who wrote the music for the musical ""A Chorus Line""?"

Answer: marvin hamlisch
}

Question: What was the name of the female that politician John Profumo had an affair with which ended his political career in 1963?

Answer:

\end{prompt}
\vspace{-5mm}
\caption{
    Example prompts for BioASQ, SQuAD, and TriviaQA.
    {\promptEmphExample Italic texts} denote 5-shot example.
}
\label{fig:appendix-prompt-uq}
\end{figure*}

\begin{figure*}[hp!]
\begin{prompt}{An answer generation prompt for GSM8K}
Given a question, please decompose it into sub-questions. For each sub-question, please answer it in a complete sentence, ending with "The answer is". When the original question is answerable, please start the subquestion with "Now we can answer the question: ".

{
\promptEmphExample

Question 1: Albert is wondering how much pizza he can eat in one day. He buys 2 large pizzas and 2 small pizzas. A large pizza has 16 slices and a small pizza has 8 slices. If he eats it all, how many pieces does he eat that day?

Question 1.1: How many slices are in one large pizza?

Answer 1.1: One large pizza has 16 slices. The answer is 16.

Question 1.2: How many slices are there in total from the large pizzas?

Answer 1.2: He buys 2 large pizzas, so 2 * 16 = 32 slices. The answer is 32.

Question 1.3: How many slices are in one small pizza?

Answer 1.3: One small pizza has 8 slices. The answer is 8.

Question 1.4: How many slices are there in total from the small pizzas?

Answer 1.4: He buys 2 small pizzas, so 2 * 8 = 16 slices. The answer is 16.

Question 1.5: Now we can answer the question: How many pieces does he eat that day?

Answer 1.5: There are 32 slices from the large pizzas and 16 slices from the small pizzas, so he eats 32 + 16 = 48 pieces that day. The answer is 48.
}

Question 2: Josh decides to try flipping a house.  He buys a house for \$80,000 and then puts in \$50,000 in repairs. This increased the value of the house by 150\%.  How much profit did he make?

Question 2.1: How much did Josh spend on the house and repairs in total?

Answer 2.1:

\end{prompt}

\begin{prompt}{An action generation prompt for GSM8K}

Given a question, please decompose it into sub-questions. For each sub-question, please answer it in a complete sentence, ending with "The answer is". When the original question is answerable, please start the subquestion with "Now we can answer the question: ".

{ \promptEmphExample

Question 1: Albert is wondering how much pizza he can eat in one day. He buys 2

large pizzas and 2 small pizzas. A large pizza has 16 slices and a small pizza has 8 slices. If he eats it all, how

many pieces does he eat that day?

Question 1.1: How many slices are in one large pizza?

Answer 1.1: One large pizza has 16 slices. The answer is 16.

Question 1.2: How many slices are there in total from the large pizzas?

Answer 1.2: He buys 2 large pizzas, so 2 * 16 = 32 slices. The answer is 32.

Question 1.3: How many slices are in one small pizza?

Answer 1.3: One small pizza has 8 slices. The answer is 8.

Question 1.4: How many slices are there in total from the small pizzas?

Answer 1.4: He buys 2 small pizzas, so 2 * 8 = 16 slices. The answer is 16.

Question 1.5: Now we can answer the question: How many pieces does he eat that day?

Answer 1.5: There are 32 slices from the large pizzas and 16 slices from the small pizzas, so he eats 32 + 16 = 48 pieces that day. The answer is 48.
}

Question 2: Josh decides to try flipping a house.  He buys a house for \$80,000 and then puts in \$50,000 in repairs.  This

increased the value of the house by 150\%.  How much profit did he make?

Question 2.1:
\end{prompt}
\begin{prompt}{A reward generation prompt for GSM8K}

Given a question and some sub-questions, determine whether the last sub-question is useful to answer the question. Output 'Yes' or 'No', and a reason.

{\promptEmphExample
Question 1: Four years ago, Kody was only half as old as Mohamed. If Mohamed is currently twice as 30 years old, how old is Kody?

Question 1.1: How old is Mohamed?

Question 1.2: How old was Mohamed four years ago?

New question 1.3: How old was Kody four years ago?

Is the new question useful? Yes. We need the answer to calculate how old is Kody now.
}

Question 2: Josh decides to try flipping a house.  He buys a house for \$80,000 and then puts in \$50,000 in repairs. This increased the value of the house by 150\%.  How much profit did he make?

New question 2.1: How much did Josh spend on the house?

Is the new question useful?

\end{prompt}
\vspace{-5mm}
\caption{
    Example prompts for GSM8K.
    {\promptEmphExample Italic texts} denote 1-shot example. The prompts are identical to those used in \citet{lee2025semantic}.
}
\label{fig:appendix-prompt-gsm}
\end{figure*}

\begin{figure*}[hp!]
\begin{prompt}{An answer generation prompt for ARC}
Given a question, please decompose it into sub-questions. For each sub-question, please answer it in a complete sentence, ending with "The answer is". When the original question is answerable, please start the subquestion with "Now we can answer the question with an option from A to D: ".

{\promptEmphExample

Question 1: Juan and LaKeisha roll a few objects down a ramp. They want to see which object rolls the farthest. What should they do so they can repeat their investigation? Options: A) Put the objects in groups, B) Change the height of the ramp, C) Choose different objects to roll, D) Record the details of the investigation.

Question 1.1: What is necessary to ensure that experimental results can be repeated?

Answer 1.1: To ensure repeatability, experimental details must be accurately recorded. The answer is to record details.

Question 1.2: What kind of information should Juan and LaKeisha record for repeatability?

Answer 1.2: They should record details like the objects used, ramp height, and surface conditions. The answer is experimental conditions.

Question 1.3: How would recording experimental details help in the investigation?

Answer 1.3: Recording details allows them to recreate the exact same conditions for reliable comparison. The answer is that it enables consistent replication.

Question 1.4: Now we can answer the question with an option from A to D: What should they do to repeat their investigation?

Answer 1.4: Record the details of the investigation. The answer is D.
}

Question 2: Which method is the safest way to watch an eclipse of the Sun? Options: A)  Turn away after two or three minutes. B)  Look at the Sun through a long telescope. C)  Cast an image through a pinhole onto a screen. D)  Blink often until your eyes get used to the light..

Question 2.1: Why should you not look directly at the Sun during an eclipse?

Answer 2.1:

\end{prompt}

\begin{prompt}{An action generation prompt for ARC}

Given a question, please decompose it into sub-questions. For each sub-question, please answer it in a complete sentence, ending with "The answer is". When the original question is answerable, please start the subquestion with "Now we can answer the question with an option from A to D: ".

{
\promptEmphExample

Question 1: Juan and LaKeisha roll a few objects down a ramp. They want to see which object rolls the farthest. What should they do so they can repeat their investigation? Options: A) Put the objects in groups, B) Change the height of the ramp, C) Choose different objects to roll, D) Record the details of the investigation.

Question 1.1: What is necessary to ensure that experimental results can be repeated?

Answer 1.1: To ensure repeatability, experimental details must be accurately recorded. The answer is to record details.

Question 1.2: What kind of information should Juan and LaKeisha record for repeatability?

Answer 1.2: They should record details like the objects used, ramp height, and surface conditions. The answer is experimental conditions.

Question 1.3: How would recording experimental details help in the investigation?

Answer 1.3: Recording details allows them to recreate the exact same conditions for reliable comparison. The answer is that it enables consistent replication.

Question 1.4: Now we can answer the question with an option from A to D: What should they do to repeat their investigation?

Answer 1.4: Record the details of the investigation. The answer is D.

}

Question 2: Which method is the safest way to watch an eclipse of the Sun? Options: A)  Turn away after two or three minutes. B)  Look at the Sun through a long telescope. C)  Cast an image through a pinhole onto a screen. D)  Blink often until your eyes get used to the light..

Question 2.1:
\end{prompt}
\begin{prompt}{A reward generation prompt for ARC}

Given a question and some sub-questions, determine whether the last sub-question is useful to answer the question. Output 'Yes' or 'No', and a reason.

{
\promptEmphExample
Question 1: How are particles in a block of iron affected when the block is melted? Options: A) The particles gain mass, B) The particles contain less energy, C) The particles move more rapidly, D) The particles increase in volume.

Question 1.1: What happens to particle energy when a solid melts?

Question 1.2: How does the movement of particles change during melting?

New question 1.3: Why does increased movement signify a phase change to liquid?

Is the new question useful? Yes, because understanding why increased movement signifies a phase change helps clarify the behavior of particles during melting.
}

Question 2: Which method is the safest way to watch an eclipse of the Sun? Options: A)  Turn away after two or three minutes. B)  Look at the Sun through a long telescope. C)  Cast an image through a pinhole onto a screen. D)  Blink often until your eyes get used to the light..

New question 2.1: Why should you not look directly at the Sun during an eclipse?

Is the new question useful?

\end{prompt}
\vspace{-5mm}
\caption{
    Example prompts for ARC.
    {\promptEmphExample Italic texts} denote 1-shot example. The prompts are identical to those used in \citet{lee2025semantic}.
}
\label{fig:appendix-prompt-arc}
\end{figure*}

\begin{figure*}[hp!]
\begin{prompt}{A pairwise clustering label generation prompt}

Given the context and a question pair, determine whether the two questions have the same meaning. Output 'Yes' or 'No', and a reason.

{
\promptEmphExample

Context: Is a Boeing 737 cost covered by Wonder Woman (2017 film) box office receipts?

Question pair: "What is the cost of a Boeing 737?" and "How much does a Boeing 737 cost?"

Answer: Yes, because both questions ask for the cost of the same airplane.

Context: Cynthia eats one serving of ice cream every night. She buys cartons of ice cream with 15 servings of ice cream per carton at a cost of \$4.00 per carton. After 60 days, how much will she spend on ice cream?

Question pair: "How many servings of ice cream does Cynthia eat in 60 days?" and "What is the total number of ice cream servings Cynthia eats?"

Answer: Yes, because both ask for the total servings of ice cream over 60 days.

Context: In a jewelers store, the price of a gold Jewell is 4/5 times as much as the price of a diamond Jewell. The cost of a silver Jewell is \$400 less than the price of gold. If a diamond Jewell is \$2000, find the total price for all three jewels.

Question pair: "How much is the price of a silver Jewell?" and "What is the price of the gold Jewell?"

Answer: No, because they refer to different types of Jewell.

}

Context: Julie is reading a 120-page book. Yesterday, she was able to read 12 pages and today, she read twice as many pages as yesterday. If she wants to read half of the remaining pages tomorrow, how many pages should she read?

Question pair: "How many pages did Julie read yesterday?" and "What is the number of pages Julie finished reading today?"

Answer:
\end{prompt}
\vspace{-5mm}
\caption{
    An example prompt for generating pairwise clustering labels.
    {\promptEmphExample Italic texts} denote 3-shot example.
}
\label{fig:appendix-prompt-label-gen}
\end{figure*}

\end{document}